\documentclass[lettersize,journal]{IEEEtran}

\usepackage{amsmath,amsfonts}
\usepackage{algorithmic}
\usepackage{algorithm}
\usepackage{array}
\usepackage[caption=false,font=normalsize,labelfont=sf,textfont=sf]{subfig}
\usepackage{textcomp}
\usepackage{stfloats}
\usepackage{url}
\usepackage{verbatim}
\usepackage{graphicx}
\usepackage{cite}

\usepackage{multirow}
\usepackage{booktabs}
\usepackage{bm}
\usepackage{dutchcal}
\usepackage{color}
\usepackage{algorithm}
\usepackage{algorithmic}
\definecolor{c1}{RGB}{0,180,255}

\begin{document}

\title{Active Domain Adaptation with Multi-level Contrastive Units for Semantic Segmentation}

\author{Hao Zhang, Ruimao Zhang,~\IEEEmembership{Member,~IEEE,}, Zhanglin Peng, Junle Wang, Yanqing Jing
\thanks{Hao Zhang and Ruimao Zhang are with The Chinese University of Hong Kong (Shenzhen), and also with Shenzhen Research Institute of Big Data, Shenzhen, China (e-mail: zhanghao1@cuhk.edu.cn and ruimao.zhang@ieee.org ).}
\thanks{Zhanglin Peng is with the Department of Computer Science, The University of Hong Kong, Hong Kong, China (e-mail: zhanglin.peng@connect.hku.hk ).}
\thanks{Junle Wang and Yanqing Jing are with Tencent, Shenzhen, China (e-mail: jljunlewang@tencent.com, and frogjing@tencent.com)}
\thanks{Corresponding Author is Ruimao Zhang}

}



\maketitle

\begin{abstract}
To further reduce the cost of semi-supervised domain adaptation (SSDA) labeling, a more effective way is to use active learning (AL) to annotate a selected subset with specific properties.
However, domain adaptation tasks are always addressed in two interactive aspects: domain transfer and the enhancement of discrimination, which requires the selected data to be both uncertain under the model and diverse in feature space.
Contrary to active learning in classification tasks, it is usually challenging to select pixels that contain both the above properties in segmentation tasks, leading to the complex design of pixel selection strategy.
To address such an issue, we propose a novel Active Domain Adaptation scheme with Multi-level Contrastive Units (ADA-MCU) for semantic image segmentation.
A simple pixel selection strategy followed with the construction of multi-level contrastive units is introduced to optimize the model for both domain adaptation and active supervised learning.
In practice, MCUs are constructed from intra-image, cross-image, and cross-domain levels by using both labeled and unlabeled pixels. 
At each level, we define contrastive losses from center-to-center and pixel-to-pixel manners, with the aim of jointly aligning the category centers and reducing outliers near the decision boundaries.
In addition, we also introduce a categories correlation matrix to implicitly describe the relationship between categories, which are used to adjust the weights of the losses for MCUs.
Extensive experimental results on standard benchmarks show that the proposed method achieves competitive performance against state-of-the-art SSDA methods with 50\% fewer labeled pixels and significantly outperforms state-of-the-art with a large margin by using the same level of annotation cost.
\end{abstract}


\begin{figure}[t]
	\begin{center}
		\includegraphics[width=\linewidth]{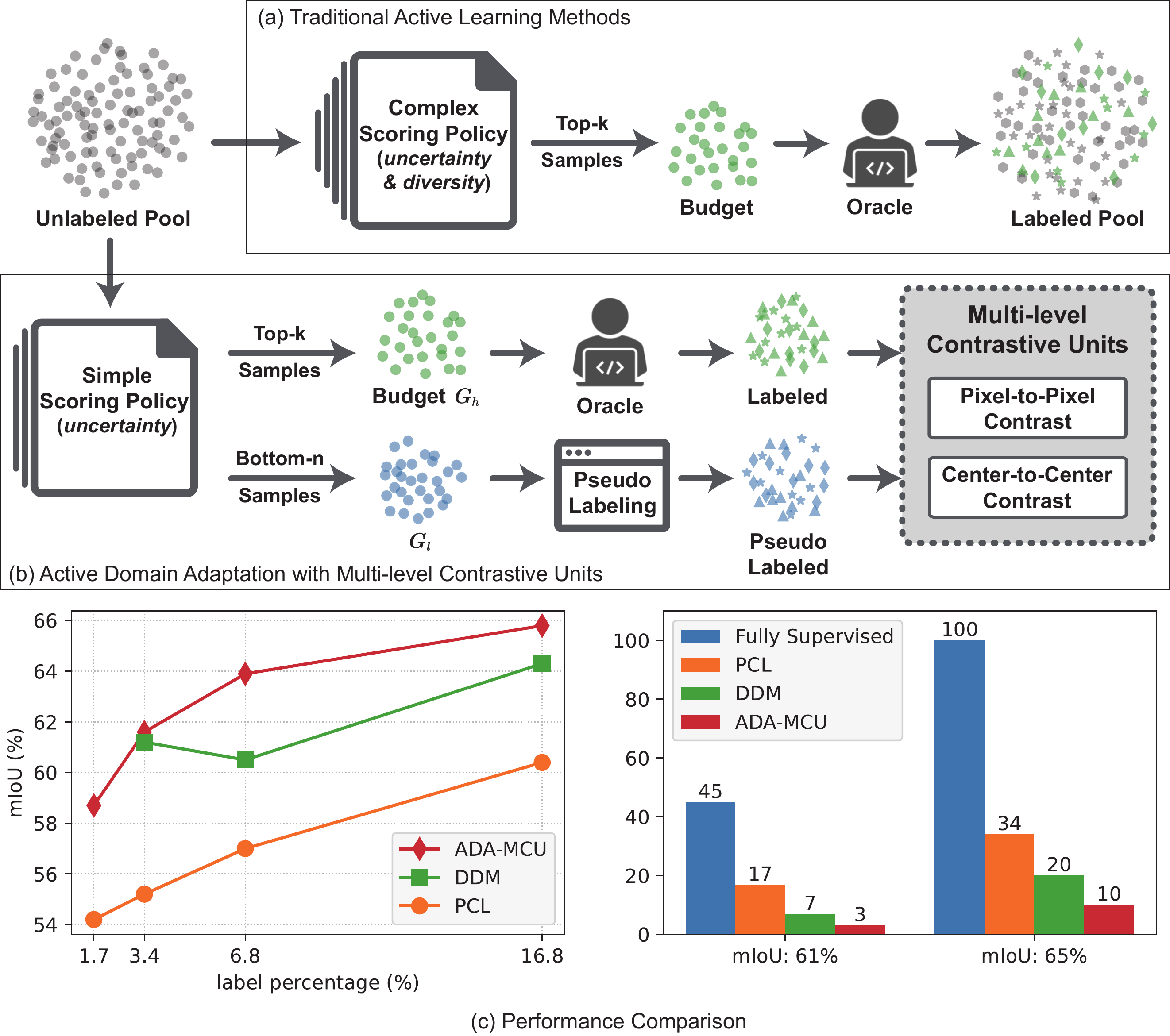}
	\end{center}
	\vspace{-10pt}
	\caption{traditional active learning methods via the complex sample selection policy for domain adaptation task and (b) the pipeline of our proposed Active Domain Adaptation with Multi-level Contrastive Units. $G_l$ and $G_h$ denote low and high uncertainty pixel groups determined by the uncertainty score. Our method abandons the complex pixel-selection policy but improves the performance of the model by constructing MCUs. It significantly outperforms current SSDA methods in both efficiency and accuracy shown in (c).} 
	\vspace{-13pt}
	\label{fig:1}
\end{figure}

\section{Introduction}
\IEEEPARstart{S}{emantic} segmentation is one of the most classic tasks in computer vision and image processing. 
The goal is to learn the semantic context in the image and automatically annotate each pixel a category label according to such learned information~\cite{C:PSPNet,J:Deeplab}.
%
%
In practice, such a task always requires excessively numerous annotations, limiting its scalability in real applications.
One way to reduce annotation cost is to leverage a large amount of virtual data that is easy to obtain labels from game engines to extend training samples (\textit{e.g.}, GTA5, SYNTHIA, Synscapes).
However, the model trained merely with virtual data performs terribly on real-world data distribution because of the domain shifts. 
Therefore, many domain adaption methods are raised in recent years to bridge the gap between label-rich virtual data and label-scarce real-life data.


One way to address the domain adaptation problem is to train the model under the semi-supervised manner (SSDA), which jointly leverages fully labeled source domain data and a subset of labeled target domain data in the training phase.
Compared to unsupervised domain adaptation (UDA), semi-supervised methods are able to significantly improve the segmentation accuracy through a small amount of annotated samples. 
Although the SSDA method makes a good balance between performance and annotation cost, there is still a big gap compared with the performance of the fully supervised approaches.
In practice, how to bridge this gap by effectively labeling data is still an open issue.

%
%
%
%
%
In the literature, active learning (AL), which aims to select a subset of samples with specific properties, is proposed to annotate the training samples in a cost-effective way ~\cite{J:lewis1994heterogeneous,J:scheffer2001active,C:jain2016active}.
Although it has extensive research~\cite{C:vezhnevets2012active} on various areas by adopting active learning, it remains a challenge to deal with domain adaptation semantic segmentation since we need to address this task by dealing with two entangled issues from the pixel level, \textit{i.e.,} \textbf{aligning domain distribution } and \textbf{ improve the model's discriminant ability}. 
In simple terms, we need to make the representative pixels (\textit{i.e.,} the ones near the category center) from the same category but different domains be closer in the feature space, and to reduce uncertainty pixels (\textit{i.e.,} the ones with low predictive confidence) of each domain by pushing them to the corresponding centers from the decision boundary.
%
%
However, it is especially challenging to select pixels that simultaneously meet the above two properties. Moreover, mining pixels being satisfied with the above properties from thousands of candidate pixels in an image through a complex scheme is also a time-consuming process.
%

%
To tackle such an issue, we propose a novel Active Domain Adaptation scheme with Multi-level Contrastive Units (ADA-MCU) for semantic segmentation.
%
As shown in Fig.~\ref{fig:1} (a) and Fig.~\ref{fig:1} (b), different from active domain adaptation for image classification, which requires a complex scoring policy to labeled specific samples for model supervision,  
%
%
we use a simple selection policy in ADA-MCU to divide each image into two subsets of pixels, (\textit{i.e.}, actively labeled ones with low confidence scores and unlabeled ones), then adopt pixels from these two subsets to construct contrastive pairs from multiple perspectives for model optimization.
%
%
%
%
%
%
%
Specifically, at the \textbf{cross-domain} level, we enforce the distribution of the feature representations in the target domain being aligned to the source domain. 
While at the \textbf{intra-image} and \textbf{cross-image} level, we enforce the instance representations belonging to the same category to be closer in the feature space and make those are belonging to different categories to be far away in both source and target domain. 
Additionally, for each contrastive level, we do the alignment in two perspectives by using two kinds of loss, \textit{i.e.}, center to center contrastive loss and pixel to pixel contrastive loss. 
The former enforces the centers of distribution to be aligned while the latter reduces the uncertainty of pixels by pulling them closer to the category center representations. 
In this way, two domains would be better aligned by employing the synergy of the above two losses.

In practice, since the misclassified pixels are highly relevant to its spatial layout, \textit{e.g.}, the {\ttfamily sidewalk} is more likely to be misclassified as the {\ttfamily road} but be rarely misclassified as {\ttfamily sky}, we further introduce a dynamic categories correlation matrix (DCCM) to model the implicit relationship between each pair of categories. 
DCCM will be updated online during the training phase, aiming to adjust the weights of contrastive losses for the MCUs. Such a categories-aware contrastive loss could further improve the discriminative feature representation learning across domains.
In this way, the domain transfer and discrimination enhancement are unified into one single framework.

The main contributions of this article can be summarized as follows,
\begin{enumerate}
    \item  We propose ADA-MCU, a novel active learning scheme, which uses a simple selection policy along with the construction of MCUs to optimize the model. 
    \item  We introduce a simple yet effective scheme to construct Multi-level Contrastive Units (MCU) to regularize model training from multiple perspectives and propose a dynamic categories correlation matrix (DCCM) to describe the implicit relationship between categories, making more effective usage of the labeled and unlabeled pixels for model training.
    \item As shown in Fig.~\ref{fig:1}~(c), extensive experiments demonstrate that our proposed method can achieve similar performance on two standard synthetic-to-real semantic segmentation benchmarks with less than $50\%$ labeled data compared with current semi-supervised domain adaptation methods, and significantly outperforms state-of-the-art with a large margin by using the same level of annotation cost.
\end{enumerate}

\section{Related Work}

\subsection{Domain Adaptation for Semantic Segmentation}
Domain adaptation task aims to explore how to transfer the knowledge that the model learned from one domain to another. In domain adaptation task, typically there are two domains, which have a certain domain shift but also share some common knowledge with each other. The domain that we train our model originally with is called source domain and the domain that we want to transfer our model to is called target domain.
According to the usage of target domain annotations, we can divide domain adaptation semantic segmentation task into two main parts: 1) unsupervised domain adaptation one termed UDA and 2) semi-supervised domain adaptation termed SSDA.
UDA is aimed at transferring the knowledge obtained from a labeled source domain to an unlabelled target domain. While SSDA aims at narrowing the gap with the fully supervised results by using a small set of labeled samples.

In recent years, some UDA methods for semantic segmentation~\cite{A:hoffman2016fcns,C:vu2019advent,A:huang2021mlan} have been proposed via adversarial training which relies on a discriminator to measure the divergence between two domains' distributions. Adversarial based methods aims at aligning the feature space of source domain and target domain by confusing the discriminator.
Image translation has also been widely used in the UDA methods~\cite{C:hoffman2018cycada,C:huang2021fsdr}. As we know the most obvious gap between source domain and target domain is the color distribution. The aligning of color distribution is a very easy but efficient way to improve the performance.
Recently some works ~\cite{C:ES, C:CaSP} proposed that class-agnostic training paradigm gives model stronger generalization ability, which also pointed out a promising way of UDA.
In addition, several methods ~\cite{C:zou2018unsupervised,C:huang2021cross} also focus on self-training that generates pseudo labels for unlabelled data in the target domain.

For the SSDA, some methods~\cite{C:wang2020alleviating,C:chen2021semi,A:huang2021semi} implemented feature alignment across domain from global and semantic level. 
For example, Chen \textit{et al.}~\cite{C:chen2021semi} proposed a framework based on dual-level domain mixing to address the differences in the amount of the labeled data between two domains. 
Huang \textit{et al.}~\cite{A:huang2021semi} aligned features by employing a few labeled target samples as anchors. However, none of those methods focus on sample selection, especially from the pixel-level perspective. 
We introduce pixel-level active learning to make more effective use of labeled pixels compared with SSDA.

\begin{figure*}[t]
	\begin{center}
		\includegraphics[width=\linewidth]{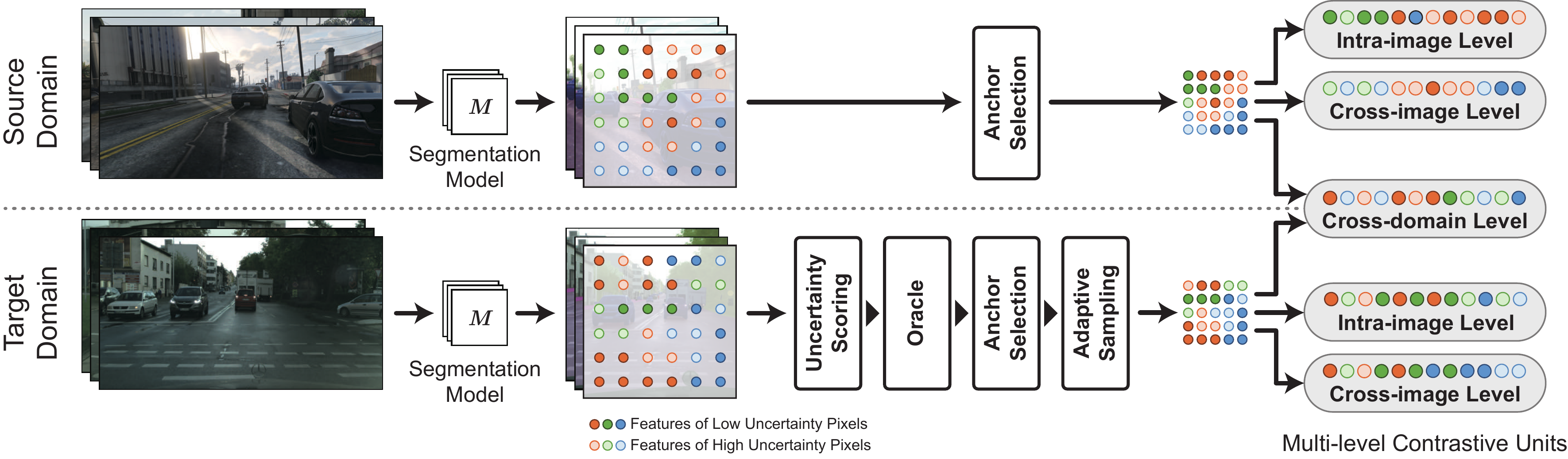}
	\end{center}
	\vspace{-10pt}
	\caption{Detailed illustration of anchors selection in source domain and target domain and construction of Multi-level Contrastive Units for model optimization.} 
	\vspace{-10pt}
	\label{fig:MCUs}
\end{figure*}

\subsection{Contrastive Learning}
Recently, contrastive learning shows very strong ability on self-supervised pretraining. Self contrastive learning~\cite{C:CPC, C:InfoMax, C:wu2018unsupervised, C:MoCo, C:SimCLR} with no ground truth labels can achieve even better result compared with supervised pretraining. 
Contrastive learning (CL) is a discriminative approach that aims at grouping similar samples
closer and diverse samples far from each other in the representation space to learn more discriminitive representations. 
%
The sample pair selection policy plays a significant role in contrastive learning. Therefore, a great number of articles focus on how to select the positive and negative sample pairs.
For image data, the common positive sample selection method is to add strong disturbances and augmentation to an image to create different views of each sample. 
At the same time, negative samples can generally be obtained directly through random sampling~\cite{C:wu2018unsupervised,A:caron2020unsupervised}. 
But for better results, some works have proposed some methods to improve performance by hard examples.
In addition, to store more negative samples during contrast computation, memory bank~\cite{C:wu2018unsupervised} or momentum updated~\cite{C:MoCo} memories are adopted.
For the semantic segmentation task, CL could also be used to do intra-domain model pre-training~\cite{A:chaitanya2020contrastive}. 
Recently, Wang \textit{et al.}~\cite{A:wang2021exploring} also propose a pixel-wise contrast scheme for semantic segmentation by exploiting inter-image pixel-to-pixel similarity to enforce global constraints on the embedding space, 
making the feature representation of pixels in each category more compact.

\subsection{Active Learning}
In the literature, active learning (AL) aims to select the most informative samples from the unlabeled dataset and hand it over to Oracle for labeling, to reduce the cost of labeling as much as possible while still maintaining performance. 
Usually, there are three major views to address AL tasks. 1) uncertainty-based methods~\cite{J:lewis1994heterogeneous,J:scheffer2001active}, uncertainty-based AL based on a common criterion that the prediction with higher confidence are more likely to be predicted correctly, while those prediction with lower confidence are more likely to be predicted wrong. Therefore, those sample with lower confident prediction contains more information, which means it's more worthwhile to be annotated. 
2) diversity-based methods~\cite{C:jain2016active,J:hoi2009semisupervised}, diversity-based AL based on another common criterion that those more representative samples from the whole dataset are more worthwhile to be annotated. In such a case, since the labeled data is more representative of the category, the model can learn a more accurate representation of the category. 
3) methods based on expected model change~\cite{C:vezhnevets2012active}. Expected model change methods are also based on a criterion, that the samples from which the model can learn more knowledge are more worthwile be annotating. Because those samples usually contain more information, which leads to more effect on the model optimization. 
Compared to these existing works in active learning, our proposed method address AL at pixel-level and we construct MCUs for assistance in order to achieve great performance without a complicated selection strategy.

\section{Methodology}
In this section, we will describe our proposed method ADA-MCU in detail. 
Firstly, we will introduce a simple pixel-level sample selection policy, dividing each image into different pixel subsets.  
Secondly,  we will present how to construct multi-level contrastive units (MCU) using labeled samples and unlabeled ones. 
At last, we will discuss how to apply contrastive loss with the dynamic category correlation to optimize the segmentation model.

\subsection{Problem Setting and Notation}
\noindent
The goal of semantic segmentation domain adaptation is to transfer the model from source domain $X_s$ to  the target domain $X_t$. In our setting, we have a fully labeled source domain $\{(x_s^n, y_s^n)\}_{n=1}^{N_s}$, indicating the $n$-th source image $x_s^n$ with the ground truth label map $y_s^n$, 
and the target domain $\{x_t^n\}_{n=1}^{N_t}$. Here $N_s$ and $N_t$ denote the number of source and target domain images, respectively.
For the pixel-level active domain adaptation, the $n$-th target image $x_t^n$ contains two subsets in pixel-level, naming active annotated pixel $x_t^{n:{(i,j)}}$ with its corresponding ground-truth label $y_t^{n:{(i,j)}}$ and unlabeled pixel ${x_t}^{n:{(\overline{i},\overline{j})}}$, where $(i,j)$ and $(\overline{i},\overline{j})$ denote the labeled and unlabeled pixel positions in the target image.
We use $M_e$ and $M_a$ to represent the number of expected labeled pixels number and already labeled pixels. And $M_t$ is the total number of pixels in the target training dataset. 
Thus the annotation rate can be presented as $R_a$ by $M_e/M_t$.

In semi-supervised domain adaptation task, we focus on narrowing domain gap with small set of annotations in target domain. 
Form task-level, we can divide SSDA task into two subtasks, (1) fully supervised DA task and (2) self-training DA task.  
In the fully supervised subtask, we use data with true labels of target domain to finetune the source-pretrained network like usual, while in the self-training subtask, we use pseudo labels that generated by pretrained model to supervise the network. 
Therefore, the annotated pixels selection and the quality of pseudo label become determining factors in SSDA. 
In this work, firstly, we use source domain training data to pretrain the network. 
After that, in target domain, we introduce active learning strategy to select the most error-prone and informative pixels to annotate and select those pixels with most convinced and accurate predictions to generate pseudo labels.
More details will be discussed in following sections.

\subsection{Active Pixel Annotation via Uncertainty Score}

\noindent 
In practice, we first use source domain images to train the segmentation network. Once the pre-trained model $\mathcal{F}$ is obtained, we use the output of $\mathcal{F}$ on both source and target images to obtain the calculate their predictive uncertainty scores.
Such selection policy is based on the uncertainty of the predictive results. 
The uncertainty score of pixel $(i, j)$ in the $n$-th image can be calculated as follows,  
\begin{equation} 
S(x^{n:(i,j)}) = E(x^{n:(i,j)}) + \gamma D_{\rm KL}(p^{n:(i,j)}||~\hat{p}^{n:(i,j)})
\label{1}
\end{equation}
where the first part  $E(\cdot)$ is the pixel-wise entropy, which is calculated by Eqn. \eqref{E}.
The second part, $D_{\rm KL}(\cdot)$ is the pixel-wise KL divergence between the predictions of the main segmentation head and  
the auxiliary head.\footnote{ The auxiliary loss is proposed in~\cite{C:PSPNet} to improve the accuracy. We leverage the outputs from both auxiliary and main segmentation heads of DeepLab v2 in most experiments to calculate KL divergence.}
$\gamma$ is a hyperparameter to control the weights of two uncertainty indicators.
$E(\cdot)$ and $D_{\rm KL}(\cdot)$ can be calculated as follows,
\begin{equation} 
E(x^{n:(i,j)}) = \frac{-1}{\log(C)} \sum_{c=1}^{C}{p_{c}^{n:(i,j)}} \log{p_{c}^{n:(i,j)}},
\label{E}
\end{equation}

\begin{equation} 
D_{\rm KL}(p~||~\hat{p}) = \sum_{c=1}^{C}{p_c}~(\log\,p_c - \log\,\hat{p_c}),
\label{KL}
\end{equation}
where $C$ denotes the total number of categories,
By calculating the entropy of the pixel and the KL divergence of two head outputs, the degree of each pixel's uncertainty is obtained for further pixel annotation.
%

%
Using the scoring function mentioned above with the threshold $\pi_{high}, \pi_{low}$, we can divide both source and target image pixels into three groups termed \textit{high}, \textit{low}, and \textit{medium} uncertainty groups respectively.
Here the higher uncertainty indicates the lower model predictive confidence.
%
%
By adjusting $\pi_{high}$, we can control the annotation rate of the target domain. 
We will discuss how to use pixels in different groups for contrastive unit construction in the next subsection.

%

\subsubsection{Category Center Generation Strategy.}
After we divide both source and target pixels of each training batch into three groups, we use those pixels from the low uncertainty group with high predictive confidence to generate the category center. 
Intuitively, high confident samples always lie in the center of category clusters, leading to a high density. According to the boundary assumption~\cite{C:chapelle2005semi}, the decision boundary should not across the high-density region of a cluster. 
In other words, high confident pixels are always reliable and can be applied as representative of the cluster. 
Therefore, the aggregation of those pixels in the low uncertainty group for each domain can be considered as the representation of each category's center.
\label{ccg}

\subsubsection{Annotation via Adaptive Sampling}
When labeling the target domain pixels, we introduce adaptive sampling (AS) to maximize the advantages of active learning strategy. 
For each training batch, after dividing each target domain image into three groups: $G_h^t$, $G_m^t$ and, $G_l^t$, we randomly select the same number of pixels in those three groups.
Then for $G_h^t$ and $G_m^t$, we give ground-truth labels to those pixels, and we use the current predictions as pseudo labels to annotate those pixels selected by AS in $G_l^t$. 
Note that the size of each group changes for each training iteration. Thus the annotated pixels in $G_h^t$, $G_m^t$ and $G_l^t$ should be dynamically sampled, and we called it \textit{annotation via adaptive sampling}. 
In practice, the size of $G_m^t$ is always larger than $G_h^t$, therefore, in other words, we give a higher sampling rate to the group with high uncertainty for labeling.

\begin{figure}[t]
	\begin{center}
		\includegraphics[width=\linewidth]{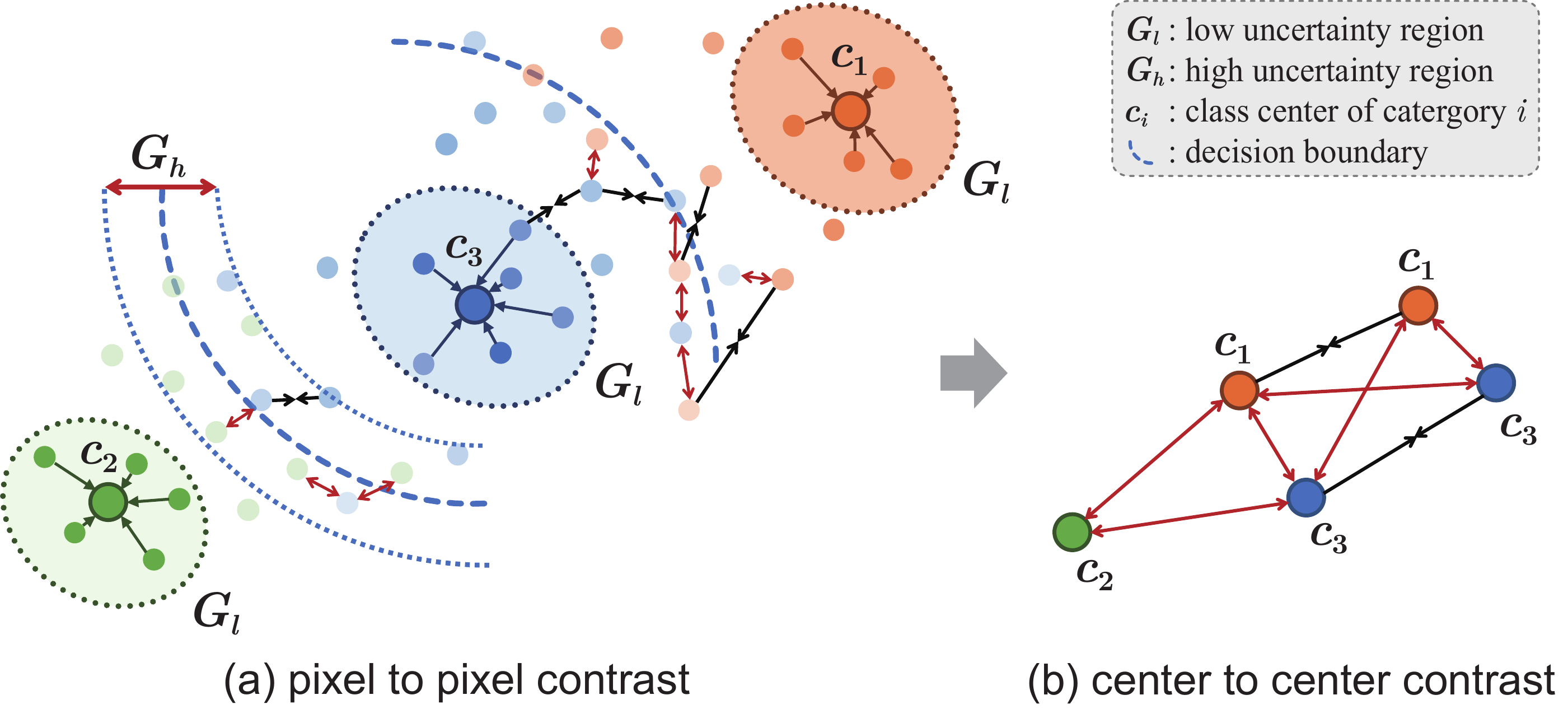}
	\end{center}
	\vspace{-10pt}
	\caption{Illustration of the construction of MCUs. We leverage both labeled data from $G_h$ (when adding adaptive sampling, we also consider $G_m$ to label) and unlabeled data from $G_l$ to construct MCUs, and in each level of MCUs we use both pixel-to-pixel and center-to-center contrast to regularize the model in both uncertainty and diversity perspectives. 
	Note that the representations can be from different pairs of images at different levels. High transparency means high uncertainty.} 
	\vspace{-10pt}
	\label{fig:overview}
\end{figure}

\subsection{Multi-level Contrastive Unit Construction}


After obtaining the active labeled target pixels, we construct the multi-level contrastive units for domain adaptation and discrimination enhancement, which is partially inspired by the pixel-wise contrastive learning for semantic segmentation ~\cite{C:wang2021exploring}.
%
%
For each batch size that contains the same number of source and target images, we construct the contrastive units in three levels: (1) intra-image level, (2) cross-image level, and (3) cross-domain level.

\subsubsection{Anchor Selection Policy.} Before constructing the contrastive units for each level, we first require to select the anchor pixels for each contrastive unit to construct. In practice, rather than mining informative samples, we follow the work~\cite{C:wang2021exploring}, determining whether it is a hard anchor through its uncertainty and whether the model predicts it correctly.
In the source domain, we compare the predictive pixels' labels and their ground truth and then select the ones which are incorrectly predicted as anchors. 
Similarly, in the target domain, we use active learning to annotate the pixels with high uncertainty.
Then the incorrectly predicted pixels in the high uncertainty group\footnote{Note that the pixel with high uncertainty does not necessarily to have the wrong predictive result.} are selected as the target anchors.

\subsubsection{Contrastive Unit Construction.} In each level, we firstly extract anchors from images using the above anchor selection policy. 
Then the pixels with the same label as the anchor are indicated as the positive samples, while the others with different labels are identified as the negative ones.
After that, we construct the contrastive units in three levels, which are illustrated in Fig.~\ref{fig:MCUs}.
For the \textbf{intra-image level}, we extract positive pixels and negative pixels from the same image according to anchors. 
While there is a problem that the independence of images leads to the loss of information across the whole dataset. 
For example, the class {\ttfamily car} and class {\ttfamily train} may never appear in the same image through the entire domain, making these two categories lack inter-actions at the intra-image level. 
In sample terms, we can't push the representations of those categories away from each other. 
In the literature, memory bank can be adopted~\cite{C:wu2018unsupervised} to address such an issue by storing the representations of other images and we could select positive samples and negative samples from such a memory bank. However, the huge memory bank always leads to huge space and time consumption.
Therefore, we introduce \textbf{cross-image level} contrast to solving this problem. For the cross-images level, we first extract anchors from one image and extract positive pixels and negatives ones from another random image in the same domain. 
In this way, any two categories could be able to have interactions with each other along with the training iterations. 
According to the above strategy, the intra-image and  the cross-image contrast units encourage the feature representation to be more discriminative in a specific domain.
In addition, we further introduce \textbf{cross-domain level} contrastive units to align source and target distribution for the domain transfer.
Different from cross-images level contrast, for cross-domain level, we extract anchors from one image in a certain domain and extract positive samples and negative ones from another image from the other domain. 
Through the synergy of three levels' contrast units, we could jointly achieve the discrimination enhancement and domain alignment in the model optimization process.

%

\subsection{Pixel-level Active Domain Adaptation with MCUs}

\subsubsection{Segmentation Loss Function.}

In target domain, we use $G_h^t$, $G_m^t$, and $G_l^t$ to denote the labeled pixels selected by adaptive sampling in high, medium, and low uncertainty groups, because there only labels for those samples selected by AL with AS in target domain.
%
%
%
Thus the active training loss by using the labeled high uncertainty pixels can be defined as,
\begin{equation} 
{\mathcal{L}}^{G_h^t,G_m^t}_{\rm seg}=\sum_{n:(i,j) \in {G_h^t,G_m^t}}{\mathcal{L}}_{\rm CE}(p^{n:(i,j)}, y^{n:(i,j)})
\label{eq:gh}
\end{equation}
where $p^{n:(i,j)}$ and $y^{n:(i,j)}$ are the prediction and ground truth of pixel in position $(i,j)$ of $n$-th target image, and ${\mathcal{L}}_{\rm CE}$ denotes the cross-entropy loss function.
On the contrary, pixels in $G_l^t$ are with low uncertainty, which means more likely to be correctly predicted.
Thus, we directly use their predictions as pseudo labels to calculate the loss,
\begin{equation} 
{\mathcal{L}}^{G_l^t}_{\rm seg}=\sum_{n:(\overline{i},\overline{j}) \in {G_l^t}}{\mathcal{L}}_{C\!E}(p^{n:(\overline{i},\overline{j})}, \widetilde{y}^{n:(\overline{i},\overline{j})}),
\label{eq:gl}
\end{equation}
%
%
where $\widetilde{y}^{n:(\overline{i},\overline{j})} = {\rm{softmax}}(p^{n:(\overline{i},\overline{j})})$. 
Thus the overall segmentation loss can be represented as follows,
\begin{equation} 
\mathcal{L}_{\rm seg}= \mathcal{L}^{G_h^t,G_m^t}_{\rm seg} + \mathcal{L}^{G_l^t}_{\rm seg} + \mathcal{L}^{G_h^s,G_m^s,G_l^s}_{\rm seg}.
\label{segloss}
\end{equation}
where $G_h^s$, $G_m^s$ and $G_l^s$ denote the pixel groups of the source domain. Note that since we have all annotations for images in source domain, we just calculate cross-entropy loss as their segmentation loss.

\subsubsection{Dynamic Categories Correlation Matrix (DCCM).} In practice, the context information and spatial layout are critical for segmentation accuracy.
For instance, class {\ttfamily road}  and {\ttfamily sidewalk} are misclassified more frequently,  while {\ttfamily sky} and {\ttfamily road} are rarely to be misclassified.
To better leverage such information, we introduce DCCM to describe the implicit relationship between any two categories.
It is also critical for the model optimization since the effect of the context information and spatial layout for the segmentation accuracy and those implicit information will be saved in DCCM.
Denote $M_k^{(c_u,c_v)}$ as the number of pixels being misclassified from category $c_u$ to $c_v$.
Let $M^{c_u}$ indicate the number of all pixels in $c_u$, then the error rate $R_k^{(c_u,c_v)}$ can be calculated as follow,
%
%
\begin{equation} 
R_k^{(c_u,c_v)} =  M_k^{(c_u,c_v)}~/~M^{c_u}
\end{equation}
And at each iteration, we could dynamically update each element of the correlation matrix $W$ by using exponential moving average as,
\begin{equation} 
\label{10}
w^{(u,v)}_{\tau} =  \beta~w^{(u,v)}_{\tau-1} + (1-\beta)~R_k^{(c_u,c_v)},
\end{equation}
where $w^{(u,v)}_{\tau}$ denotes the correlation coefficient of $u$-th and $v$-th categories at the iteration $\tau$.
Then DCCM will further guide multi-level contrastive units by adjusting the weight of contrastive loss for each MCU.


\subsubsection{Contrastive Loss Function.}

In each level of MCUs, we define loss function in two perspectives.
On the one hand, we introduce pixel-to-pixel ({\ttfamily p2p}) contrastive loss based on the labeled pixels to reduce their uncertainty by pulling the same class pixel being close and pushing different class samples being apart. 
%
%
The loss function is defined based on InfoNCE~\cite{C:oord2018representation}, modified by using the weights in DCCM.
\begin{equation} 
\mathcal{L}^{\rm p2p}_{\rm con} = 
\frac{1}{|P_p|}\sum_{p^+ \in P_p} -\log~\mathcal{H}_p,
\end{equation}

\begin{equation} 
\label{12}
\mathcal{H}_p = \frac{\exp(w^{(u,v)} \ p\cdot p^+/\lambda)}{\exp(w^{(u,v)}\  p\cdot p^+/\lambda)+\sum_{p^- \in N_p } \exp(w^{(u,v)}\   p\cdot p^-/\lambda)},
\end{equation}
where $\cdot$ denotes dot multiplication of two vectors with the scalar as the output. $P_p$ and  $N_p$ denote pixel-wise embedding collections of the positive and negative samples.


On the other hand, we introduce class-to-class ({\ttfamily c2c}) contrastive loss by using the category centers introduced in the former section, which are most representative for each category.
Then we define {\ttfamily c2c} contrastive loss in the same form with the {\ttfamily p2p},

\begin{equation} 
\label{13}
\mathcal{L}^{\rm c2c}_{\rm con} =
\frac{1}{|P_c|}\sum_{c^+ \in P_c} -\log~\mathcal{H}_c,
\end{equation}
where $P_c$ and  $N_c$ denote class-wise embedding collections of the positive and negative samples, and $\mathcal{H}_c$ has the same form as Eqn.~\ref{12} but uses class-wise anchors instead of pixel-wise ones. 
Using two kinds of contrastive losses as regular terms, we can encourage the features from the same category to be closer and from different categories to be further. And the DCCM can increase the weight between the categories which are more likely to be misclassified so as to guiding model optimization.  
In this way, the total loss of the proposed ADA-MCU can be presented as $\mathcal{L}_{\rm total} = \mathcal{L}_{\rm seg} +\mathcal{L}_{\rm con}$, 
where 
$\mathcal{L}_{\rm con}$ is the sum of $(\mathcal{L}_{\rm con}^{\rm p2p} + \mathcal{L}_{\rm con}^{\rm c2c})$ from three levels.


\section{Experiment}
\subsection{Benchmarks}
We firstly evaluate our proposed method by using two standard large-scale segmentation benchmarks for domain adaptation, GTA5-to-Cityscapes, and SYNTHIA-to-Cityscapes. 
In addition, we further validate our proposed method by using a new synthetic dataset SYNSCAPES\cite{C:ros2016synthia} as the source domain to prove the generalization of our proposed method.
Following the previous method~\cite{C:MME}, we apply $19$ classes domain adaptation for the GTA5-to-Cityscapes and SYNSCAPES-to-Cityscapes evaluation, and $13$ classes for the SYNTHIA-to-Cityscapes evaluation.

\begin{enumerate}
    \item \textit{Cityscapes dataset.} It consists of $5000$ real-world urban scene images with $2048 \times 1024$ resolution and dense-pixel annotation. There are $2975$ images, which are split for training, $500$ images, which are split for validation, and $1525$ images, which are split for testing. Cityscapes dataset annotates have $33$ categories and $19$ of them are used for training and evaluation. %
    \item \textit{GTA5 dataset.} It contains $24966$ densely annotated images, synthesized by a game engine with a resolution of $1914 \times 1052$. The real label of its ground floor is consistent with the urban landscape.
    \item \textit{SYNTHIA dataset.} It is a large synthetic data set consisting of photorealistic frames rendered from virtual cities. In the experiment, we used the Synchy-Rand cityscape setting to adapt. It contains $9,400$ images with a resolution of $1280 \times 760$, which are labeled in $16$ categories. Like GTA5 dataset, its annotations are automatically generated and compatible with the Cityscape dataset.
    \item \textit{SYNSCAPES dataset.} It is also a synthetic dataset created by using photorealistic rendering techniques. It consists of $25,000$ images at $1440 \times 720$ resolution, with 33 category-dense annotations, of which only $19$ are used. Again, its annotations are compatible with Cityscape. Syncscapes is closer to Cityscape than GTA5 and Synscapes
\end{enumerate}

\subsection{Experiment Setting and Implementation}
\noindent
We conduct extensive experiments and report mean Intersection-over-Union (mIoU) compared with existing domain adaptation methods.
%
Without special notes, all of the methods employ Deeplab v2~\cite{J:Deeplab} as the basic model, which utilizes a pre-trained ResNet-101~\cite{C:he2016deep} on ImageNet as backbone network. 
To measure the uncertainty, we calculate the KL divergence in Eqn.~\eqref{KL} by using multi-level outputs coming from both {\ttfamily conv4} and {\ttfamily conv5} feature maps.
All experiments are run on a single Tesla V100 GPU with 32 GB of memory. All the models are trained by the Stochastic Gradient Descent (SGD) optimizer with an initial learning rate of $2.5\times10^{-4}$ and decreasing with the polynomial annealing procedure with the power of $0.9$, the momentum 0.9 and weight decay $10^{-4}$. The learning rate is decreased with the polynomial annealing procedure with power of 0.9.

For coarse aligning two domains, we follow the method~\cite{C:vu2019advent} to use CycleGAN~\cite{C:zhu2017unpaired} to translate source domain images in order to reduce the visual difference between source and target domain images.
Before the domain adaptation procedure, we use translated source domain data to train the model first. Then in each iteration, we randomly select $4$ images, contains $2$ from the source domain and $2$ from the target domain. 
For \textbf{source domain images}, we feed them to the model and get the predictions and pixel-wise features. After that, we use predictions and labels to calculate cross-entropy loss, execute anchor selection and calculate intra-image and cross-image contrastive loss. 
While for \textbf{target domain images}, since we don't have any annotations at the beginning, we directly calculate the pixel-wise uncertainty score of the model output (after {\ttfamily softmax} function) and divide the batch of images into three groups according to the uncertainty, where $\gamma$ is set to $0.5$ in Eqn.~\eqref{1}. 
We ask the Oracle to label the high uncertainty group, and use the predictions as pseudo labels for pixels in low uncertainty group. 
Then, we can calculate cross-entropy loss and intra-image and cross-image contrastive loss similar to soure domain. 
Furthermore, we also calculate cross-domain contrastive loss using both source and target domain data.

\begin{algorithm}[]
\caption{The training pipeline of proposed ADA-MCU}
\label{alg:algorithm1}
\textbf{Require}: Labeled source dataset $D_s = \{(x_s^n, y_s^n)\}^{Ns}_{n=1}$,  unlabeled target dataset $D_t = \{x_t^n\}^{Nt}_{n=1}$,
segmentation network $\mathcal{F}$.\\
\textbf{Parameter}: Parameters of network: $\theta$, number of training epochs: $T$, decay rate: $r$ (defult to be 0.9), budget of expected annotated pixels in target domain $M_e$, number of already labeled pixels: $M_a$, dynamic correlation category matrix $W$. \\
%
%
\textbf{Procedure}:
\begin{algorithmic}[1] 
\STATE Use CycleGAN to translate images in $D_s$ to having a similar appearance as target images following the same setting in work~\cite{vu2019advent}.
\STATE Use $D_s$ to train the segmentation network $\mathcal{F}$.
\STATE Set epoch variable $i=0$.
\WHILE{$i<= N_e$}
\STATE Randomly load two images from $D_s$, and two images from $D_t$, 
\IF{$M_a < M_e/2$}
\STATE Update $\pi_{high}$ = $\pi_{high} \times r$ and $G_h$. Select and annotate pixels in $G_h$ follow the anchor selection policy.
\STATE Update $M_a$ as the number of selected pixels.
\ELSIF {$M_a == M_e/2$}
\STATE Stop selecting pixels from $G_h$ but select and annotate $M_e/2$ of pixels in $G_m$.
\ELSE 
\STATE Sample $M_e/2$ pixels from $G_l$ follow the anchor selection policy and using their predictions from  $\mathcal{F}$ as the pseudo labels.
\ENDIF
\STATE Update $W$ according to the Eqn. (8) in the main article. 
\STATE Calculate segmentation loss and multi-level contrastive losses.

\STATE Update the network parameters $\theta$ and $i=i+1$.
\ENDWHILE
\STATE \textbf{return} $\theta$
\end{algorithmic}
\end{algorithm}

\subsection{Technique Details about Adaptive Sampling } 
There exist several simple methods to control the annotation percentage to a specific level that we expect.
Firstly, we can achieve it at the image-level by specifying that select the same percentage of pixels to label in each image so that the percentage of labeling on the entire dataset are controlled at the same ratio. 
While the disadvantage of this method is that some images may contain more outliers that need to be labeled. 
A straightforward way to solve such a problem is to use the memory to store all of the pixels, sorted by their uncertainty score. 
However, it's difficult to achieve in practice since it requires a huge memory cost, and it's hard to be progressive because of the time-consuming.

To tackle the above issue, in practice, we introduce a sample yet effective way to adjust $\pi_{high}$,
which is shown in Algorithm~\ref{alg:algorithm1}. 
In the beginning, we initialize $\pi_{high}$ with a large value.
Then, in each training epoch, if the number of annotated pixels $M_a$ is still lower than the expected quantity, we decrease $\pi_{high}$ until $M_a$ achieves the number of expected labeled high uncertainty $M_e$. 
For example, assume we set $M_e$ equals $5\%$ of total pixels in the target domain. Firstly we initialize $\pi_{high}$ with a big number, and as a result, only around $1\%$ of samples are selected as $G_h$. 
Then we reduce $\pi_{high}$, which will lead to the increment in the size of $G_h$ \footnote{Note that the size of $G_h$ doesn't equal $M_a$, because we only sample half of the pixels in $G_h$ uniformly to annotate, so the size of $G_h$ equals $2M_a$}. 
When the $M_a$ equals to $M_e/2$, which is about $2.5\%$ in our assumption, we stop selecting samples from $G_h$.
Then we directly select another $M_e/2$ samples uniformly from $G_m$ for annotation. After that, we stop giving annotations and only apply predictions to those samples in $G_l$ as their pseudo labels. 
%

\begin{table}[]
\small
\caption{Experimental results on GTA5-to-Cityscapes compared with current SSDA, semi-supervised learning (SSL) methods. $19$-class mIoU (\%) scores are reported on Cityscapes validation set by using $1.7\%$, $3.4\%$, $6.8\%$, $16.8\%$ labeled pixels from whole dataset. Note that $1.7\%$ pixels of $2795$ images are at the same pixel number of $50$ images.}
\label{table1}
\begin{center}
\begin{tabular}{c|l|cclll}
\bottomrule[2pt]
\multirow{2}{*}{Type} & \multicolumn{1}{c|}{\multirow{2}{*}{Method}} & \multicolumn{5}{c}{Annotation Percentage (\%)} \\
 & \multicolumn{1}{c|}{}  & \multicolumn{1}{l}{1.7} & 3.4 & 6.8 & 16.8 \\ \hline
\multirow{1}{*}{Supervised} & \multicolumn{1}{c|}{Image-wise} & - & 41.9 & 47.7 & 55.5 \\ \hline
\multirow{2}{*}{SSL}  & \multicolumn{1}{c|}{CutMix (\textcolor{c1}{bmvc20})}  & - & 50.8 & 54.8 & 61.7 \\
 & \multicolumn{1}{c|}{DST-CBC (\textcolor{c1}{arxiv20})}  & - & 48.7 & 54.1 & 60.6 \\ \hline
\multirow{8}{*}{SSDA} 
 & \multicolumn{1}{c|}{MME (\textcolor{c1}{cvpr19})} &  -& 52.6& 54.4& 57.6 \\
 & \multicolumn{1}{c|}{MinEnt (\textcolor{c1}{cvpr19})} & 47.5& 49.0& 52.0& 55.3 \\
 & \multicolumn{1}{c|}{AdvEnt (\textcolor{c1}{cvpr19})} & 44.9& 46.9& 50.2& 55.4 \\
 & \multicolumn{1}{c|}{ASS (\textcolor{c1}{cvpr20})} &  50.1& 54.2& 56.0& 60.2 \\
 & \multicolumn{1}{c|}{FDA (\textcolor{c1}{cvpr20})} & 53.1& 54.1& 56.2& 59.2 \\
 & \multicolumn{1}{c|}{PCL (\textcolor{c1}{arxiv21})} & 54.2& 55.2& 57.0& 60.4 \\
  & \multicolumn{1}{c|}{DDM (\textcolor{c1}{cvpr21})}  & - & 61.2 & 60.5 & 64.3 \\\hline
 & \multicolumn{1}{c|}{Ours}  & \textbf{58.7} & \textbf{61.6} & \textbf{63.9} &  \textbf{65.8}\\ \bottomrule[2pt]
\end{tabular}
\end{center}
\vspace{-12pt}
\end{table}

\subsection{Comparison with State-Of-The-Art SSDA Methods}
\noindent
As presented in Table~\ref{table1} and Table~\ref{table2}, we compare the proposed method with two SSL methods, \textit{i.e.} CutMix~\cite{C:CutMix}, DST-CBC~\cite{C:DST}, and seven SSDA methods, \textit{i.e.}  MME~\cite{C:MME}, ASS~\cite{C:ASS}, MinEnt~\cite{C:vu2019advent}, AdvEnt~\cite{C:vu2019advent}, FDA~\cite{C:FDA}, PCL~\cite{C:PCL}, DDM~\cite{C:chen2021semi}, in different percentage of annotation: $1.7\%$, $3.4\%$, $6.8\%$, $16.8\%$. 
As expected, compared with those methods, our purposed method has achieved a significant accuracy (mIoU) improvement.

\begin{figure*}[t]
	\begin{center}
		\includegraphics[width=\linewidth]{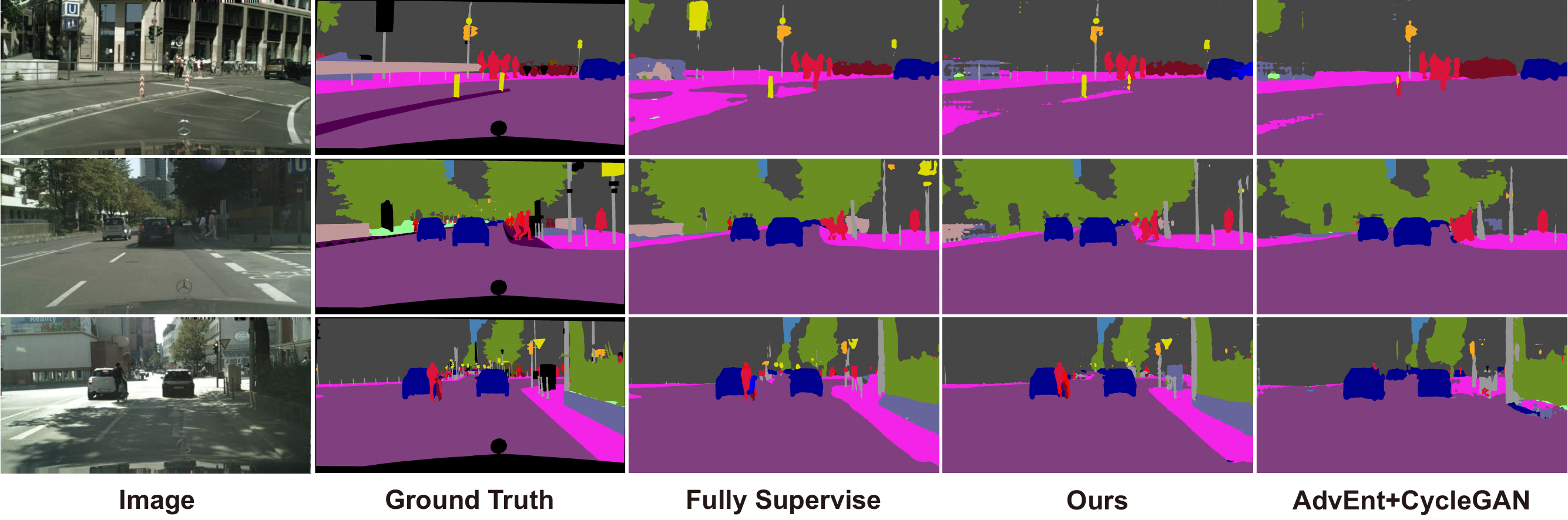}
	\end{center}
	\vspace{-10pt}
	\caption{Visualization of the segmentation results. Ours stands for our proposed method with \textcolor[rgb]{1,0,0}{20\%} active annotations.} 
	\vspace{5pt}
	\label{fig:segmentation_result}
\end{figure*}

From the Table~\ref{table1}, we can clearly see that our method can achieve comparable results with only about $50\%$ of the annotations compared with other methods.
Even compared with the state of art SSDA method DDM, we can still achieve similar performance using $30\%$ fewer annotations. 
In addition, our proposed method can achieve similar performance with the fully supervised method using only $16.8\%$ annotation of the whole target set. 
The visualization of the segmentation results of fully supervised method, UDA method~\cite{C:vu2019advent} and our proposed method with 20\% annotations are shown in Fig.~\ref{fig:segmentation_result}. We can clearly see that our proposed scheme with $20\%$ annotations can obviously improve the effect of some critical small areas, \textit{e.g.} {\ttfamily sign}, {\ttfamily rider} and {\ttfamily person}.
%

%
To further evaluate the generalization of our proposed method, we conduct experiments to adapt from the SYNSCAPES dataset to the Cityscapes dataset.
As shown in Table~\ref{synscapes}, we compared our method with the following settings: 1) source only, which only using SYNSCAPES dataset to train the model with full supervision and test it on Cityscapes dataset, 
2) AdvEnt~\cite{C:vu2019advent}, which is a well-known unsupervised domain adaptation method, and 
3) DDM~\cite{C:chen2021semi}, which is one of the state of art semi-supervised domain adaptation method. In practice, it additionally uses $6.8\%$ annotated images from Cityscapes dataset. 
At the bottom, we show the result of our proposed method, which uses $5\%$ annotations from the Cityscapes dataset. The experiment shows that our full model has a $0.8\%$ gain with fewer annotations compared with DDM.

\begin{table}[]
\small
\caption{Experimental results on SYNTHIA-to-Cityscapes compared with current SSDA and semi-supervised learning (SSL) methods. $13$-class mIoU (\%) scores are reported on Cityscapes validation set.}
\label{table2}
\begin{center}
\begin{tabular}{c|l|cclll}
\bottomrule[2pt]
\multirow{2}{*}{Type} & \multicolumn{1}{c|}{\multirow{2}{*}{Method}} & \multicolumn{5}{c}{Annotation Percentage (\%)} \\
 & \multicolumn{1}{c|}{}  & \multicolumn{1}{l}{1.7} & 3.4 & 6.8& 16.8 \\ \hline

\multirow{1}{*}{Supervised} & \multicolumn{1}{c|}{Image-wise} & - & 53.0 & 58.9 & 61.0 \\ \hline
\multirow{2}{*}{SSL}  & \multicolumn{1}{c|}{CutMix (\textcolor{c1}{bmvc20})}  & - & 61.3 & 66.7 & 71.7 \\
 & \multicolumn{1}{c|}{DST-CBC (\textcolor{c1}{arxiv20})}  & - & 59.7 & 64.3 & 68.9 \\ \hline
\multirow{8}{*}{SSDA} 
 & \multicolumn{1}{c|}{MME (\textcolor{c1}{cvpr19})} &  -& 59.6& 63.2& 66.7 \\
 & \multicolumn{1}{c|}{MinEnt (\textcolor{c1}{cvpr19})} & 52.9& 56.4& 57.9& 62.5 \\
 & \multicolumn{1}{c|}{AdvEnt (\textcolor{c1}{cvpr19})} & 51.4& 55.2& 59.6& 62.6 \\
 & \multicolumn{1}{c|}{ASS (\textcolor{c1}{cvpr20})} &  60.7& 62.1& 64.8& 69.8 \\
 & \multicolumn{1}{c|}{FDA (\textcolor{c1}{cvpr20})} & 58.5& 62.0& 64.4& 66.8 \\
 & \multicolumn{1}{c|}{PCL (\textcolor{c1}{arxiv21})} & 61.2& 63.4& 65.2& 70.3 \\
 & \multicolumn{1}{c|}{DDM (\textcolor{c1}{cvpr21})}  & - & 68.4 & 69.8 & 71.7 \\\hline
 & \multicolumn{1}{c|}{Ours}  & \textbf{66.2} & \textbf{69.1} & \textbf{70.6} &  \textbf{73.1}\\ \bottomrule[2pt]
\end{tabular}
\end{center}
\vspace{-12pt}
\end{table}

\begin{table*}[t]
\small
\caption{Experimental results on SYNSCAPES-to-Cityscapes, which using \textcolor[rgb]{1,0,0}{$5\%$} of cityscapes dataset annotations. Compared with source only, AdvEnt~\cite{vu2019advent},and state of art semi-supervised method~\cite{C:chen2021semi}, which using \textcolor[rgb]{1,0,0}{$6.8\%$} of cityscapes dataset annotations.}
\label{synscapes}
\begin{center}
\setlength\tabcolsep{3pt}
\begin{tabular}{clllllllllllllllllll|c}
\bottomrule[2pt]
\multicolumn{21}{c}{SYNSCAPES to Cityscapes} \\ \hline
\multicolumn{1}{l}{} & \rotatebox{90}{road} & \rotatebox{90}{sidewalk} & \rotatebox{90}{building} & \rotatebox{90}{wall} & \rotatebox{90}{fence} & \rotatebox{90}{pole} & \rotatebox{90}{light} & \rotatebox{90}{sign} & \rotatebox{90}{vege} & \rotatebox{90}{terrace} & \rotatebox{90}{sky} & \rotatebox{90}{person} & \rotatebox{90}{rider} & \rotatebox{90}{car} & \rotatebox{90}{truck} & \rotatebox{90}{bus} & \rotatebox{90}{train} & \rotatebox{90}{motor} & \rotatebox{90}{bike} & mIoU \\ \hline

\multicolumn{1}{l|}{$\rm Source Only$} & 79.5 & 39.6 & 75.9 & 26.2 & 25.6 & 34.7 & 34.6 & 39.3 & 82.3 & 18.7 & 84.2 & 58.2 & 37.1 & 70.4 & 19.6 & 15.2 & 5.2 & 22.0 & 54.5 & 43.3\\ \hline
\multicolumn{1}{l|}{$\rm AdvEnt$} & 93.9 & 59.4 & 84.8 & 28.0 & 26.4 & 38.0 & 43.2 & 43.3 & 85.9 & 28.4 & 88.6 & 60.9 & 35.6 & 86.9 & 31.7 & 45.7 & 24.8 & 24.6 & 56.6 & 51.9\\ \hline
\multicolumn{1}{l|}{$\rm DDM \textcolor[rgb]{1,0,0}{(6.8\%)}$} & -- & -- & -- & -- & -- & -- & -- & -- & -- & -- & -- & -- & -- & -- & -- & -- & -- & -- & -- & 62.5\\ \hline
\multicolumn{1}{l|}{$\rm Full~Model\textcolor[rgb]{1,0,0}{(5\%)}$} & 96.6 & 76.0 & 88.3 & 43.4 & 40.8 & 45.0 & 46.0 & 63.4 & 89.0 & 52.9 & 90.4 & 67.9 & 44.0 & 90.5 & 59.7 & 73.0 & 47.6 & 23.6 & 61.6 & \textbf{63.3}  \\ \bottomrule[2pt]
\end{tabular}
\end{center}
\end{table*}

\begin{table*}[t]
\small
\caption{Experimental results on GTA5-to-Cityscapes compared with current Active learning domain adaptation (ADA) method~\cite{C:ning2021multi} with \textcolor[rgb]{1,0,0}{$5\%$} annotations. The segmentation network used in the above experiment is DeepLab v3+~\cite{chen2018encoder}, which utilizes a pre-trained ResNet-101 on ImageNet as the backbone.}
\label{ada}
\begin{center}
\setlength\tabcolsep{3pt}
\begin{tabular}{clllllllllllllllllll|c}
\bottomrule[2pt]
\multicolumn{21}{c}{GTA5 to Cityscapes (DeepLab v3+)} \\ \hline
\multicolumn{1}{l}{} & \rotatebox{90}{road} & \rotatebox{90}{sidewalk} & \rotatebox{90}{building} & \rotatebox{90}{wall} & \rotatebox{90}{fence} & \rotatebox{90}{pole} & \rotatebox{90}{light} & \rotatebox{90}{sign} & \rotatebox{90}{vege} & \rotatebox{90}{terrace} & \rotatebox{90}{sky} & \rotatebox{90}{person} & \rotatebox{90}{rider} & \rotatebox{90}{car} & \rotatebox{90}{truck} & \rotatebox{90}{bus} & \rotatebox{90}{train} & \rotatebox{90}{motor} & \rotatebox{90}{bike} & mIoU \\ \hline

\multicolumn{1}{l|}{$\rm MADA$ (\textcolor{c1}{iccv21})} & 95.1 & 69.8 & 88.5 & 43.3 & 48.7 & 45.7 & 53.3 & 59.2 & 89.1 & 46.7 & 91.5 & 73.9 & 50.1 & 91.2 & 60.6 & 56.9 & 48.4 & 51.6 & 68.7 & 64.9\\ \hline
\multicolumn{1}{l|}{$\rm Full~Model$} & \textbf{97.3} & \textbf{78.5} & \textbf{88.7} & \textbf{50.8} & 44.3 & \textbf{49.6} & 49.5 & \textbf{64.1} & \textbf{89.3} & \textbf{55.9} & \textbf{91.8} & 68.7 & 37.5 & \textbf{91.6} & \textbf{65.9} & \textbf{74.6} & \textbf{58.6} & 41.5 & 65.5 & \textbf{66.5} \\ \bottomrule[2pt]
\end{tabular}
\end{center}
\end{table*}

\subsection{Comparison with State-Of-The-Art ADA Methods} 

Recently, one article termed Multi-Anchor Active Domain Adaptation (MADA)~\cite{C:ning2021multi}, which is about active domain adaptation for semantic segmentation, has been accepted by ICCV2021 as the oral representation.
The authors claim that it is the first study to adopt active learning to assist the domain adaptation regarding the semantic segmentation tasks, which adopts multiple anchors obtained via clustering-based method to characterize the feature distribution of the source-domain and multi-anchor soft-alignment loss to push the features of the target samples towards multiple anchors leading to better latent representation.
Since the DeepLab v3+~\cite{chen2018encoder} is adopted as the backbone network of this method, we follow their experiment setting in this subsection for fair comparison.
%


Different from our proposed \textbf{\textit{pixel-level}} group partition regarding the uncertainty, such a method still adopts an \textbf{\textit{image-level}} scheme to conduct active target sample selection against source anchors.
As presented in Table~\ref{ada}, we compare our method with this active domain adaptation (ADA) method, MADA~\cite{C:ning2021multi}. 
%
%
We set $M_e = 5\%$ since MADA also selects $5\%$ target-domain samples as active samples for their experiments. 
The other settings are the same as the former experiments. 
The experiment shows that our proposed method outperform MADA by a large margin, \textit{i.e.}, $1.6\%$ mIoU, which demonstrates the proposed method could take little annotation workload but brings large performance gain compared with the recent state-of-the-art method.

\subsection{Ablation Study}

\subsubsection{Effectiveness of AL with Adaptive Sampling.}
In active domain adaptation task, the active learning selection policy is especially important. 
%
Firstly, we want to investigate the effectiveness of our proposed active learning strategy.
As shown in in Table.~\ref{table1}, \textbf{UDA} is the performance of an unsupervised domain method AdvEnt with cycleGAN~\cite{C:vu2019advent}. 
\textbf{RBA} denotes the performance of a pixel-level Region-based active learning method proposed in ~\cite{C:kasarla2019region}. 
\textbf{AL(w/o~AS)} indicates the degradation model which only uses the pixels from the high uncertainty group (\textit{i.e.} pixels with manual labeling) and all of the pixels from the low uncertainty group (\textit{i.e.} pixels with pseudo labels) to optimize the model.
%
%
%
According to the results shown in Table.~\ref{table1}, we surprisingly find out that in our work, merely adding an active learning strategy may lead to performance degradation, \textit{e.g.} $43.2\%$ compared with $46.3\%$.
One acceptable reason is that our proposed scheme not only uses labeled samples but also uses some of the unlabeled ones with pseudo labels to supervise the network, and the proportion of selected labeled pixels in different images varies greatly.
For instance, if we set the annotation percentage for the whole dataset to be $10\%$, some of the images may get $2\%$ or less labeled data (\textit{e.g.} images with simple scenes) yet more than $50\%$ pseudo label to supervise the model together. 
In such case, the loss calculated by those ground truth labels would be overwhelmed by the loss calculated by the pseudo labels, especially when the annotation budget is extremely limited.
After we introduce adaptive sampling (AS) to our pixel-level active learning scheme, we get a satisfactory result ($64.3\%$ mIoU) shown by \textbf{AL(w~AS)} in Table.\ref{table1}. 
Such result also outperforms \textbf{RBA} by $2.5\%$, showing the superiority of our proposed active selection strategy.

\begin{table}[]
\small
\caption{Experiments setting of ablation study.}
\label{table3}
\begin{center}
\begin{tabular}{llllll}
\bottomrule[2pt]
 & AS & MCU-$L_i$ & MCU-$L_d$ & DCCM\\
 \hline
$\rm AL(w/o~AS)$ &  &  &  &  &\\
$\rm AL(w~AS)$  & \checkmark &  &  &  &\\
$\rm AL(w~MCU_i)$  & \checkmark & \checkmark &  &  &\\
$\rm AL(w~MCU_d)$  & \checkmark & \checkmark & \checkmark &  &\\
$\rm Full~Model$  & \checkmark & \checkmark & \checkmark & \checkmark &\\
\bottomrule[2pt]
& AS & MCU$\mathtt -p2p$ & MCU$\mathtt -c2c$ & DCCM\\
 \hline
$\rm AL(w~MCU_i)$  & \checkmark & \checkmark &  &  &\\
$\rm AL(w~MCU_d)$  & \checkmark & \checkmark & \checkmark &  &\\
$\rm Full~Model$  & \checkmark & \checkmark & \checkmark & \checkmark &\\
\bottomrule[2pt]
\end{tabular}
\vspace{0pt}
\end{center}
\end{table}

\begin{table*}[]
\small
\caption{Evaluation of different components of proposed method on GTA5-to-Cityscapes, with \textcolor[rgb]{1,0,0}{$20\%$} labeled pixels except UDA.}
\label{table0}
\begin{center}
\setlength\tabcolsep{3pt}
\begin{tabular}{clllllllllllllllllll|c}
\bottomrule[2pt]
\multicolumn{21}{c}{GTA5 to Cityscapes} \\ \hline
\multicolumn{1}{l}{} & \rotatebox{90}{road} & \rotatebox{90}{sidewalk} & \rotatebox{90}{building} & \rotatebox{90}{wall} & \rotatebox{90}{fence} & \rotatebox{90}{pole} & \rotatebox{90}{light} & \rotatebox{90}{sign} & \rotatebox{90}{vege} & \rotatebox{90}{terrace} & \rotatebox{90}{sky} & \rotatebox{90}{person} & \rotatebox{90}{rider} & \rotatebox{90}{car} & \rotatebox{90}{truck} & \rotatebox{90}{bus} & \rotatebox{90}{train} & \rotatebox{90}{motor} & \rotatebox{90}{bike} & mIoU \\ \hline

\multicolumn{1}{l|}{$\rm UDA$ (\textcolor{c1}{cvpr19})} & 92.4 & 52.7 & 83.8 & 32.4 & 24.1 & 30.7 & 33.2 & 25.7 & 83.7 & 35.1 & 85.1 & 58.2 & 27.4 & 85.5 & 37.1 & 41.9 & 2.1 & 25.2 & 22.6 & 46.3  \\
\multicolumn{1}{l|}{$\rm RBA$ (\textcolor{c1}{wacv19})} & -- & -- & -- & -- & -- & -- & -- & -- & -- & -- & -- & -- & -- & -- & -- & -- & -- & -- & -- & 61.8\\

\multicolumn{1}{l|}{$\rm AL(w/o~AS)$} & 90.4 & 34.9 & 82.3 & 30.0 & 23.4 & 27.4 & 31.9 & 21.9 & 84.0 & 38.5 & 77.8 & 58.4 & 25.0 & 84.8 & 26.9 & 34.9 & 1.5 & 27.6 & 19.8 & 43.2\\
\multicolumn{1}{l|}{$\rm AL(w~AS)$} & 96.4 & 75.7 & 86.8 & 40.3 & 42.0 & 47.4 & 46.1 & 65.4 & 87.9 & 44.0 & 84.3 & 68.6 & 44.9 & 91.5 & 66.7 & 72.6 & 53.9 & 41.9 & 64.7 & 64.3  \\
\multicolumn{1}{l|}{$\rm AL(w~MCU_i)$} & 96.8 & 76.5 & 86.8 & 40.5 & 43.2 & 47.4 & 48.5 & 66.3 & 88.6 & 50.7 & 80.7 & 69.4 & 48.4 & 91.7 & 67.4 & 73.2 & 54.2 & 45.6 & 66.6 & 65.5\\
\multicolumn{1}{l|}{$\rm AL(w~MCU_d)$} & 97.1 & 77.4 & 87.8 & 42.1 & 43.9 & 48.1 & 47.4 & 65.3 & 87.4 & 55.1 & 82.9 & 72.1 & 49.1 & 91.2 & 70.4 & 73.1 & 55.3 & 45.7 & 66.3 & 66.1\\ \hline 
\multicolumn{1}{l|}{$\rm AL(w~{\mathtt p2p})$} & 97.3 & 76.6 & 87.5 & 44.6 & 43.7 & 47.6 & 47.2 & 66.1 & 87.4 & 46.1 & 83.9 & 71.3 & 48.1 & 91.2 & 67.3 & 72.9 & 55.5 & 42.6 & 65.2 & 65.4\\  
\multicolumn{1}{l|}{$\rm AL(w~{\mathtt p2p\!+\!c2c})$} & 97.1 & 77.4 & 87.8 & 42.1 & 43.9 & 48.1 & 47.4 & 65.3 & 87.4 & 55.1 & 82.9 & 72.1 & 49.1 & 91.2 & 70.4 & 73.1 & 55.3 & 45.7 & 66.3 & 66.1\\ \hline
\multicolumn{1}{l|}{$\rm Full~Model$} & \textbf{97.2} & \textbf{78.3} & \textbf{88.4} & \textbf{46.0} & 42.9 & \textbf{48.5} & \textbf{48.6} & \textbf{66.5} & \textbf{89.2} & 54.9 & \textbf{89.3} & 70.3 & \textbf{49.7} & \textbf{92.1} & \textbf{70.9} & 72.2 & 49.0 & \textbf{46.4} & \textbf{67.0} & \textbf{66.7} \\ \bottomrule[2pt]
\end{tabular}
\end{center}
\vspace{-0pt}
\end{table*}

\noindent
\begin{figure}[t]
	\begin{center}
		\includegraphics[width=\linewidth]{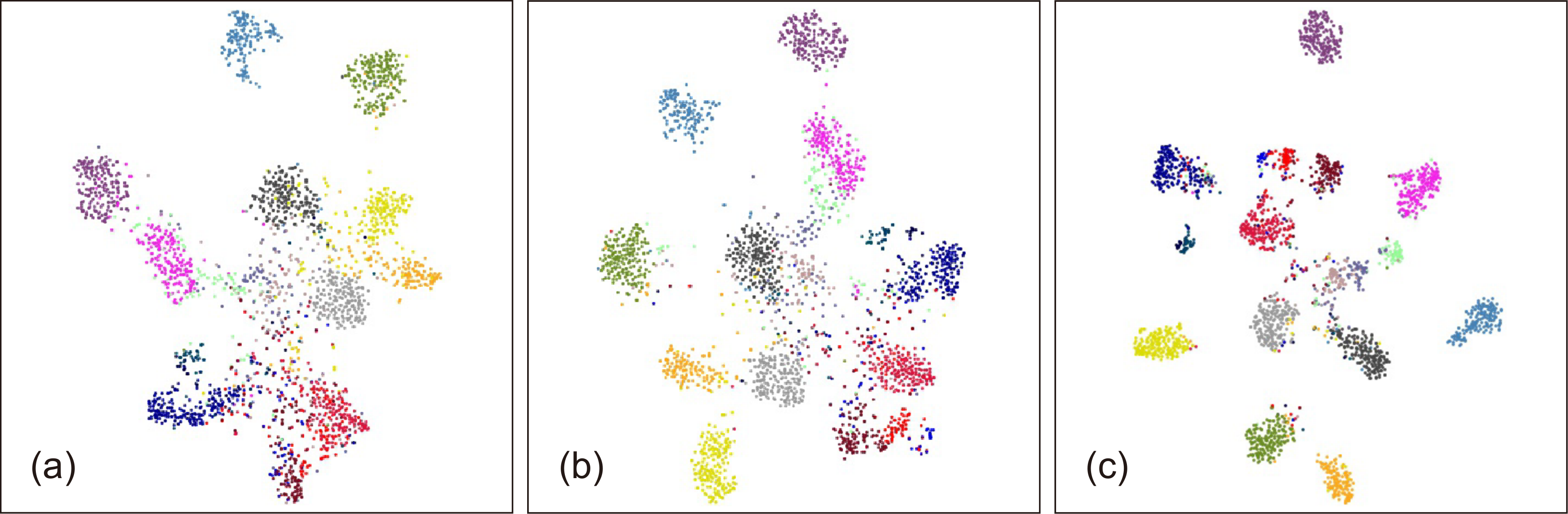}
	\end{center}
	\vspace{-10pt}
	\caption{Visualization of features learned (a) by UDA (AdvEnt), (b) using AL selection policy with adaptive sampling  and (c) adding MCUs. $19$ classes are reported on Cityscapes validation set.} 
	\vspace{-10pt}
	\label{fig:feature}
\end{figure}

\subsubsection{Effectiveness of MCUs.}
To investigate the effectiveness of MCUs, we further use the following two degradation models.
(1) \textbf{AL(w~MCU$_i$)} indicates the model that we employ active learning strategy with adaptive sampling to select annotated pixels and construct image-level contrastive units. 
(2) \textbf{AL(w~MCU$_d$)} denotes the model that we construct both image-level and domain-level MCUs. 
At last, \textbf{Full Model} denotes our entire schemes, which further adds dynamic categories correlation matrix (DCCM) to the setting.
%

As shown in Table.~\ref{table0}, firstly, by considering intra-image and cross-image relations in each domain, \textbf{AL(w~MCU$_i$)} achieves a substantial performance gain, \textit{i.e.} $1.2\%$, compared with \textbf{AL(w~AS)}.
Secondly, we investigate the effectiveness of domain-level contrast. \textbf{AL(w~MCU$_d$)} in Table.~\ref{table0} shows a further performance gain of $0.6\%$.  
Finally, we investigate the effectiveness of DCCM by comparing the results with and without DCCM. After we leverage DCCM to adjust the weight inside MCUs, we obtain a performance gain of $0.6\%$ (\textit{i.e.}, \textbf{Full Model}).
We could find that the performance of {\ttfamily road} and {\ttfamily sidewalk}, which are always being misclassified, is improved.

As shown in Fig.~\ref{fig:feature}, active learning based supervision in (b) can make the features of each category being separated, while adding MCUs in (c) can enforce features from the same category to be more compact and further separated from the features from other categories. 
Additionally, the features from some categories (\textit{e.g.}, the wathet cluster at the top of Fig.~\ref{fig:feature}~(b)) are already separated from the others. 
Thus we hope the model paying less attention to such categories, but paying more attention to those that are not separated well (\textit{e.g.}, the red and the blue clusters). 
In practice, combining MCUs with DCCM could well address such an issue. 
%
%
Just as shown in Fig.~\ref{fig:feature} (c), the red cluster is well separated from the blue one.
The above visualization results have further demonstrated the effectiveness of proposed MCUs.


\subsubsection{Effectiveness of {\ttfamily c2c}~/~{\ttfamily p2p} Contrast} 

As mentioned before, the \textbf{AL(w~{AS})} in Table~\ref{table0} is about the experiment that we implement an active learning selection strategy with proposed adaptive sampling. 
\textbf{AL(w~{\ttfamily p2p})} denotes the method that only calculates pixel-to-pixel contrastive units in three levels based on the above experiment setting. 
\textbf{AL(w~{\ttfamily p2p+c2c})} indicates the method that we apply both {\ttfamily p2p} and {\ttfamily c2c} to construct the contrastive units. 
Note that \textbf{AL(w~{\ttfamily p2p+c2c})} has the same setting as \textbf{AL(w~}\bm{${\rm MCU_d}$}\textbf{)}, and the DCCM is not adopted in such a case.

According to Table.~\ref{table0}, we can see that using pixel-to-pixel contrast, \textit{i.e.}, \textbf{AL(w~{\ttfamily p2p})}, can improve the performance by $1.1\%$ compared with \textbf{AL(w~AS)}. 
And further adding center-to-center contrast, \textit{i.e.} \textbf{AL(w~{\ttfamily p2p+c2c})}, can further improve performance by $0.7\%$.
Note that the method only uses the center-to-center contrast but without considering the pixel-to-pixel one could not achieve desired performance.
According to the \cite{wu2018unsupervised} and \cite{he2020momentum}, a large set of negative samples is critical for contrastive representation learning.
As the number of center representations in both source and target domain is limited, thus there are not enough negative samples to be used to calculate the contrastive losses, leading to poor accuracy.
Thus we have not listed the corresponding results in Table~\ref{table0}.

Additionally, we also investigate the effectiveness of pixel-to-center contrastive loss. 
However, the result of applying both {\ttfamily p2p} contrast and {\ttfamily p2c} contrast shows limited improvement compared with only adding {\ttfamily p2p} contrast. 
After careful analysis, we think it's because of the special form of InfoNCE Loss~\cite{C:oord2018representation} defined as follows,
\begin{equation} 
\label{1}
\mathcal{L}^{NCE}_{I} = -\log~\frac{\exp(v\cdot v^+/\lambda)}{\exp(v\cdot v^+/\lambda)+\sum_{v^- \in N } \exp(v\cdot v^-/\lambda)},
\end{equation}
where $v$, $v^-$, and $v^+$ denote the anchor, negative sample, and positive sample. The operation $\cdot$ denote the vector dot product. $N$ denotes the set of negative samples.  And $\lambda>0$ is a temperature hyper-parameter.
When we calculate {\ttfamily p2p} contrastive loss using pixels from all three groups (\textit{i.e.}, $G_l$, $G_m$ and $G_h$), every anchor has negative/positive samples not only from $G_h, G_m$, but also from $G_l$. 
\textit{ The partial loss from {\ttfamily p2p}, which is calculated by a specific anchor and its corresponding samples from $G_l$, seem to have a similar effort to the {\ttfamily p2c} contrastive loss.} 
This is because the center representation of each category is aggregated from the pixels from $G_l$.
As mentioned before,  we use those pixels from $G_l$ (\textit{i.e.} the low uncertainty group) with high predictive confidence to generate the category center.  
Intuitively, high confident samples always lie in the center of category clusters, \textit{leading to a high density}.
Thus the generated category centers would have very similar representations to the corresponding pixels in $G_l$ that is collected from various images.
It contributes limited to further improve the performance.

\section{Conclusion} 
In this paper, we propose ADA-MCU, a novel active learning method, which uses a simple selection policy along with the construction of MCUs to optimize the segmentation model.
As shown in Fig.~\ref{fig:1}, such a scheme abandons the complex sample selection policy in previous methods, leading to a more efficient active supervised training process.
%
%
To the best of our knowledge, this work is the first study to conduct pixel-level annotation-based active domain adaptation for semantic image segmentation. 
%
%
%
The multi-level contrastive units (MCU), together with dynamic categories correlation matrix (DCCM), are carefully designed for efficient active supervised model training, 
leading to many appealing benefits. (1) It enables the models to learn more compact feature representation for each category.
(2) It could employ fewer annotations ($16.8\%$) to achieve comparable performance with fully supervised method ($65.3\%$ mIoU). 
(3) It is effective for dealing with boundaries and small objects.
Future work will combine the proposed scheme with more powerful architecture, \textit{e.g.,} vision transformer, to explore more challenge tasks, such as panoptic segmentation.

\end{document}